%% file: main.tex
\documentclass[lettersize,journal]{IEEEtran}
\usepackage{lipsum} 
\usepackage{amsmath,amsfonts}
\usepackage{algorithmic}
\usepackage{algorithm}
\usepackage{array}
\usepackage{textcomp}
\usepackage{stfloats}
\usepackage{url}
\usepackage{verbatim}
\usepackage{graphicx}
\usepackage[detect-all]{siunitx}
\usepackage{cite}
\usepackage{comment}
\usepackage{multirow}
\usepackage{xcolor}
\usepackage{amssymb}
\usepackage{subcaption}
\usepackage{booktabs}
\usepackage[hidelinks]{hyperref}
\usepackage{etoolbox,xspace}
\usepackage{cuted}
\usepackage{capt-of}
\usepackage{enumitem}

\newcommand{\methodName}{HaHeAE\xspace}

\def\shownotes{0}

\ifnum\shownotes=1
\newcommand\zhiming[1]{\textcolor{cyan}{Zhiming: #1}}

\newcommand\todo[1]{\textcolor{magenta}{#1}}
\newcommand\syn[1]{\textcolor{green}{Syn: #1}}
\newcommand\hl[1]{\textcolor{red}{#1}}
\newcommand\andreas[1]{\textcolor{orange}{Andreas: #1}}
\newcommand\daniel[1]{\textcolor{blue}{Daniel: #1}}
\else 
\newcommand\andreas[1]{}
\newcommand\syn[1]{}
\newcommand\zhiming[1]{}
\newcommand\daniel[1]{}
\newcommand\hl[1]{#1}
\newcommand\todo[1]{}
\fi

\usepackage{graphicx}
\graphicspath{{figs/}{figures/}{pictures/}{images/}{assets/}{./}} 
\usepackage{etoolbox}

\begin{document}
\title{\methodName: Learning Generalisable Joint Representations of Human Hand and Head Movements in Extended Reality}
\author{Zhiming Hu, Guanhua Zhang, Zheming Yin, Daniel H\"{a}ufle, Syn Schmitt, Andreas Bulling \IEEEcompsocitemizethanks{\IEEEcompsocthanksitem 
Zhiming Hu, Guanhua Zhang, Zheming Yin, Syn Schmitt, and Andreas Bulling are with the University of Stuttgart, Germany. \protect E-mail: \{zhiming.hu@vis.uni-stuttgart.de, guanhua.zhang@vis.uni-stuttgart.de, st178328@stud.uni-stuttgart.de, schmitt@simtech.uni-stuttgart.de, andreas.bulling@vis.uni-stuttgart.de\}. Daniel H\"{a}ufle is with the University of T\"{u}bingen, Germany. E-mail: \{daniel.haeufle@uni-tuebingen.de\}.
Daniel H\"{a}ufle, Syn Schmitt, and Andreas Bulling are with the Center for Bionic Intelligence Tübingen Stuttgart (BITS), Germany. Zhiming Hu is the corresponding author.
}

\thanks{Manuscript May 13, 2025.}}



\maketitle
\input{sections/abstract}
\input{sections/introduction}
\input{sections/related_work}
\input{sections/method}
\input{sections/experiments}
\input{sections/applications}
\input{sections/discussion}
\input{sections/conclusion}

\section*{Acknowledgement}
This work was funded, in part, by the Baden-W\"{u}rttemberg Stiftung in the scope of the AUTONOMOUS ROBOTICS project \textit{iAssistADL} granted to Syn Schmitt and Daniel Häufle.

{
    \bibliographystyle{IEEEtran}
    \bibliography{references.bib}
}


\section{Biography Section}
\vspace{-30pt}

\begin{IEEEbiography}[{\includegraphics[width=1in,height=1.25in,clip,keepaspectratio]{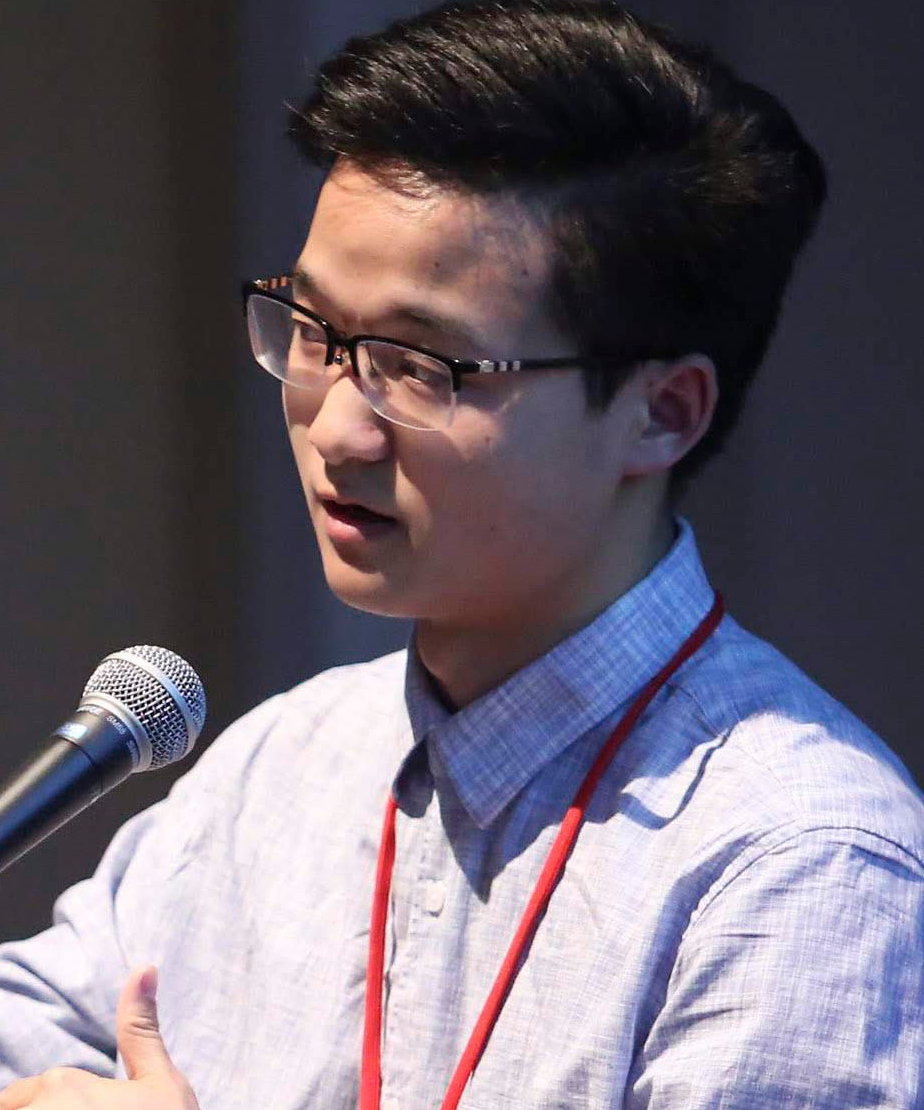}}]{Zhiming~Hu} 
is a post-doctoral researcher at the University of Stuttgart, Germany. 
He obtained his PhD in Computer Software and Theory from Peking University, China, in 2022 and his Bachelor's in Optical Engineering from Beijing Institute of Technology, China, in 2017. His research interests include virtual reality, human-computer interaction, eye tracking, and human-centred artificial intelligence.
\end{IEEEbiography}

\vspace{-40pt}

\begin{IEEEbiography}[{\includegraphics[width=1in,height=1.25in,clip,keepaspectratio]{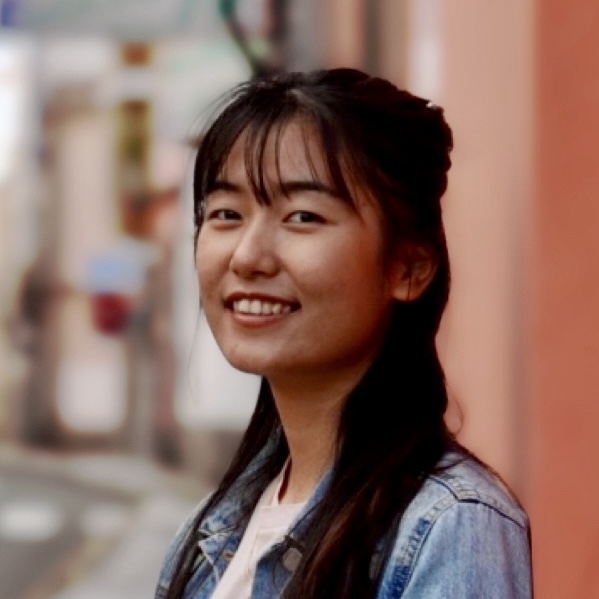}}]{Guanhua~Zhang} 
is an applied scientist at Zalando, Germany.
She received her PhD from the University of Stuttgart, Germany, in 2025, MSc. from Tsinghua University, China, in 2020 and her BSc. from Beijing University of Posts and Telecommunications, China, in 2017, all in Computer Science.
Her research interests include user modelling, human-computer interaction and machine learning.
This work was done during her PhD.
\end{IEEEbiography}

\vspace{-40pt}

\begin{IEEEbiography}[{\includegraphics[width=1in,height=1.25in,clip,keepaspectratio]{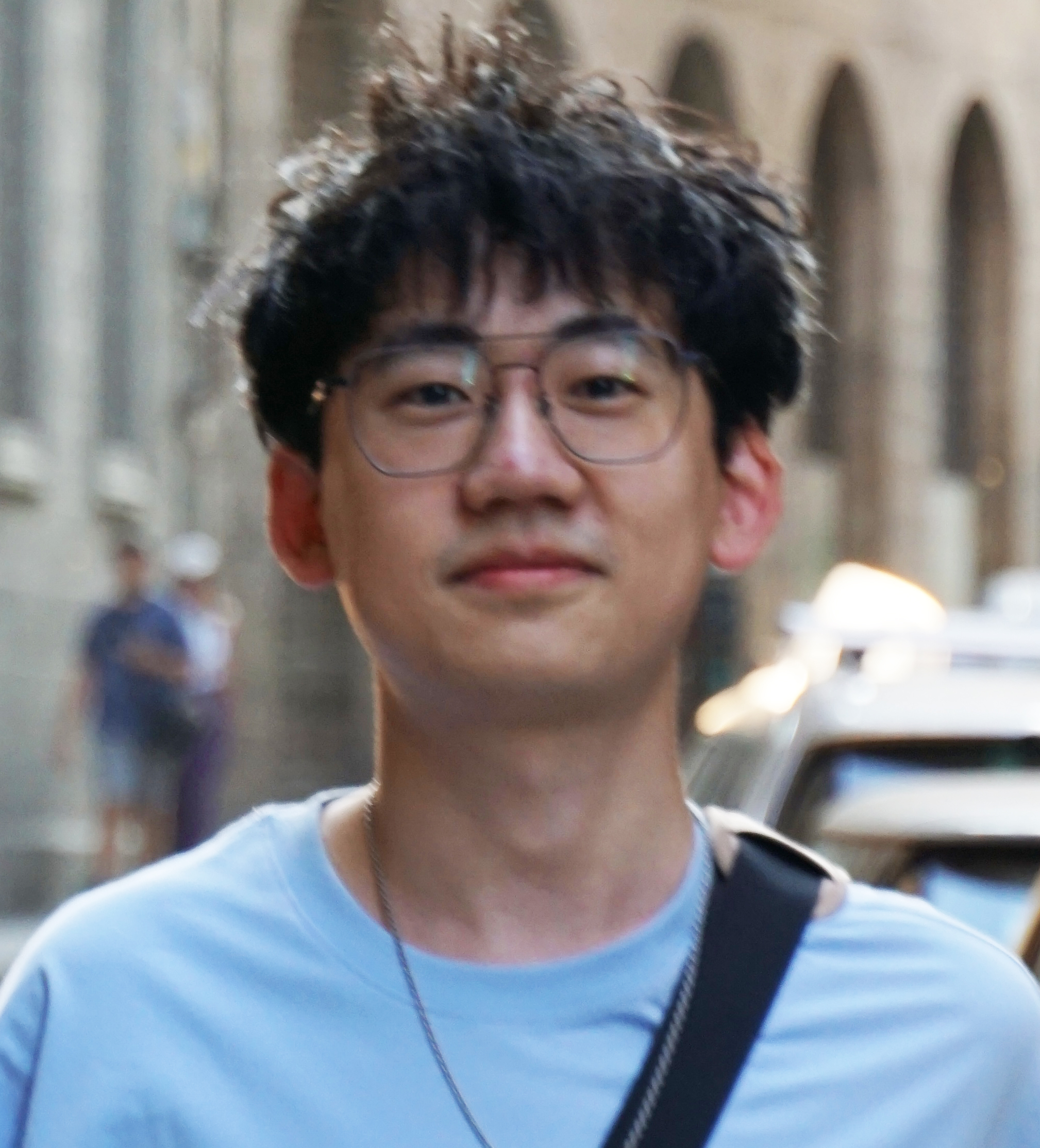}}]{Zheming~Yin} 
is a master student studying Elektromobility at the University of Stuttgart, Germany. He received his Bachelor’s degree in Automotive Engineering from Northeastern University, China, in 2021. His research interests include autonomous driving, human-computer interaction, indoor localisation, and computer vision.
\end{IEEEbiography}

\vspace{-40pt}

\begin{IEEEbiography}[{\includegraphics[width=1in,height=1.25in,clip,keepaspectratio]{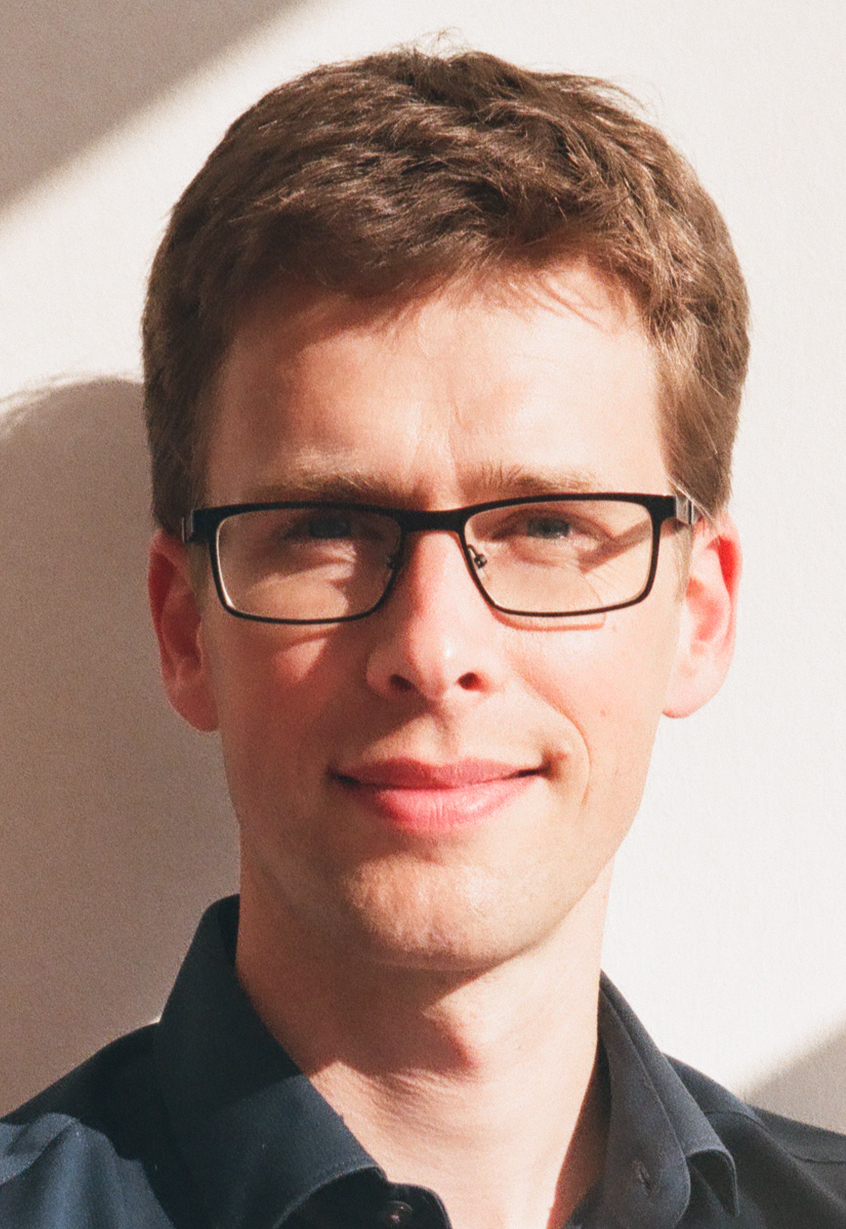}}]{Daniel~H\"{a}ufle}
is Full Professor of Neuromechanics and Rehabilitation Robotics at the University of Tübingen, Germany. He received his MSc. in Physics from the University of Jena, Germany, in 2009 and his PhD in Theoretical Physics from the University of Stuttgart, Germany, in 2012. Daniel H\"{a}ufle was a Fulbright Fellow at the Carnegie-Mellon University's Robotics Institute and was a Professor of Scientific Computing at Heidelberg University, Germany. His research interests include biomechanics, neuromechanics, and human-robot interaction.
\end{IEEEbiography}

\vspace{-40pt}

\begin{IEEEbiography}[{\includegraphics[width=1in,height=1.25in,clip,keepaspectratio]{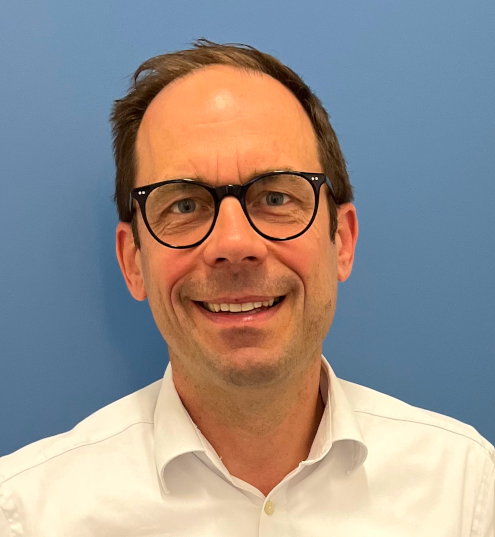}}]{Syn~Schmitt}
is Full Professor of Biomechanics and Biorobotics at the University of Stuttgart, Germany, where he directs the research group "Computational Biophysics and Biorobotics". He received his MSc. in Physics from the University of Stuttgart, Germany, in 2003 and his PhD in Theoretical Physics from the University of T\"{u}bingen, Germany, in 2006. His research interests include biomechanics, neuromechanics, and biorobotics.
\end{IEEEbiography}

\vspace{-40pt}

\begin{IEEEbiography}[{\includegraphics[width=1in,height=1.25in,clip,keepaspectratio]{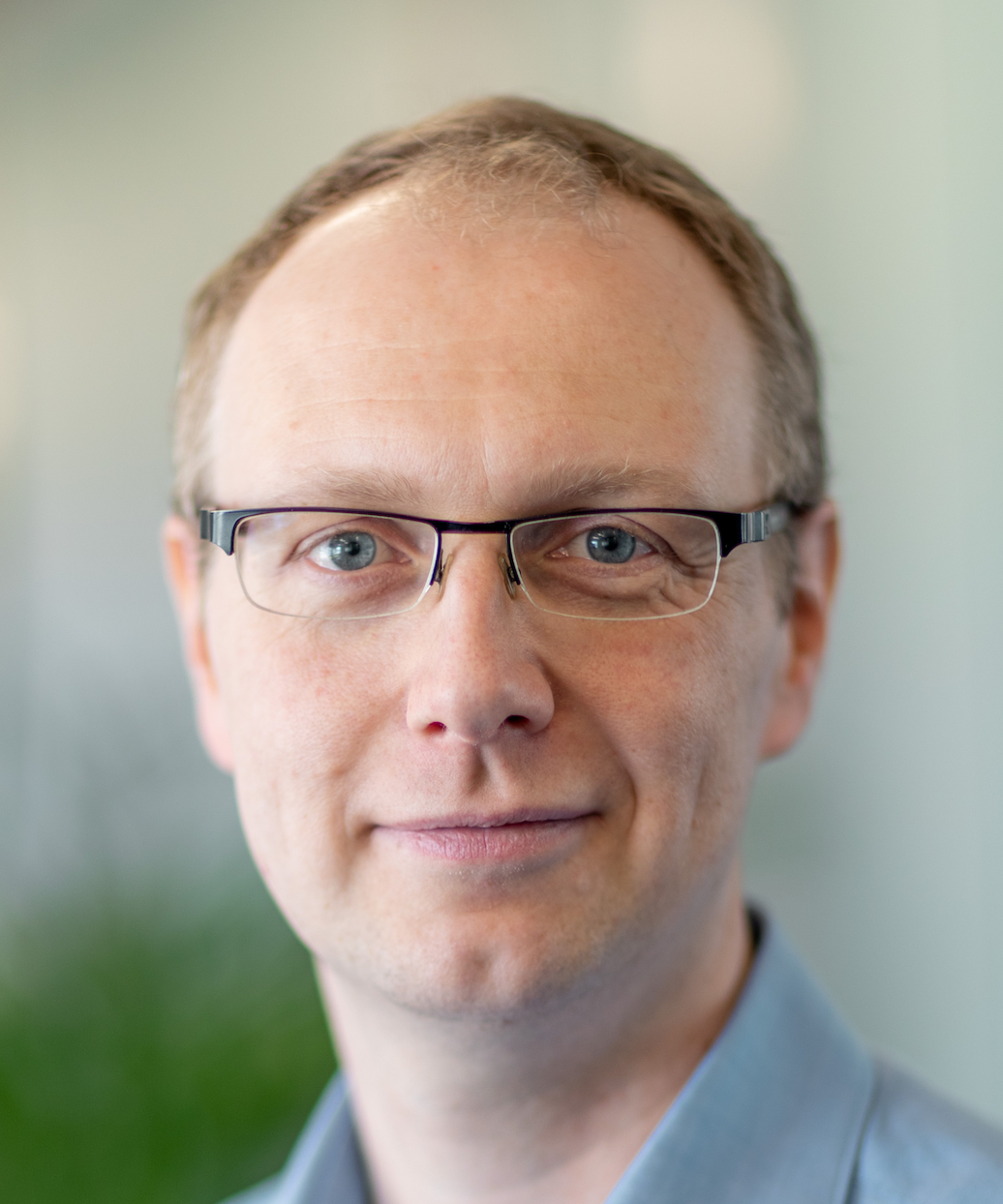}}]{Andreas~Bulling}
is Full Professor of Computer Science at the University of Stuttgart, Germany, where he directs the research group "Collaborative Artificial Intelligence". He received his MSc. in Computer Science from the Karlsruhe Institute of Technology, Germany, in 2006 and his PhD in Information Technology and Electrical Engineering from ETH Zurich, Switzerland, in 2010. Before, Andreas Bulling was a Feodor Lynen and Marie Curie Research Fellow at the University of Cambridge, UK, and a Senior Researcher at the Max Planck Institute for Informatics, Germany. His research interests include computer vision, machine learning, and socially interactive agents.
\end{IEEEbiography}

\vfill
\end{document}

%% file: sections/abstract.tex
\begin{abstract}
Human hand and head movements are the most pervasive input modalities in extended reality (XR) and are significant for a wide range of applications.
However, prior works on hand and head modelling in XR only explored a single modality or focused on specific applications.
We present \methodName~-- a novel self-supervised method for learning generalisable joint representations of hand and head movements in XR.
At the core of our method is an autoencoder (AE) that uses a graph convolutional network-based semantic encoder and a diffusion-based stochastic encoder to learn the joint semantic and stochastic representations of hand-head movements.
It also features a diffusion-based decoder to reconstruct the original signals.
Through extensive evaluations on three public XR datasets, we show that our method 1) significantly outperforms commonly used self-supervised methods by up to 74.1\% in terms of reconstruction quality and is generalisable across users, activities, and XR environments, 2) enables new applications, including interpretable hand-head cluster identification and variable hand-head movement generation, and 3) can serve as an effective feature extractor for downstream tasks.
Together, these results demonstrate the effectiveness of our method and underline the potential of self-supervised methods for jointly modelling hand-head behaviours in extended reality.
\end{abstract}

\begin{IEEEkeywords}
Hand movement, head movement, extended reality, representation learning, graph convolutional network, diffusion models
\end{IEEEkeywords}

%% file: sections/introduction.tex
\section{Introduction}

Human hand and head movements are among the most widely used input modalities in extended reality (XR) and are, as such, crucial to various XR applications.
This includes, for example, interaction target prediction (predict a user's target object in interactive virtual environments based on their hand motion features)~\cite{belardinelli2022intention}, redirected walking (redirect a user's walking path based on their head orientation)~\cite{gandrud2016predicting}, reducing cybersickness that users frequently suffer from using intentional head movements~\cite{lin2022Intentional}, user identification (identify a user amongst others based on their hand and head movements)~\cite{nair2024berkeley}, or activity recognition (recognise user activities from their head movements)~\cite{hu2022ehtask}.

Prior works on modelling hand and head movements in XR typically only focused on a single modality~\cite{gamage2021so, hu2022ehtask, hu2019sgaze, hu2020dgaze}, thus neglecting the fact that hand and head movements are closely coordinated with each other in almost all activities.
For example, when reaching for an object, the user's hands approach the specific position of the target object while their head is oriented at the object to aid the hand movement.
As such, jointly modelling hand and head movements in XR has significant potential for understanding human behaviours and developing future human-aware intelligent XR systems~\cite{hu2022ehtask, nair2024berkeley}.
In addition, existing methods on modelling hand-head movements usually focused on extracting features that are geared to a specific XR application~\cite{belardinelli2022intention, gandrud2016predicting, nair2024berkeley, hu2022ehtask}.
However, designing application-specific features is laborious, requires expert domain knowledge, and limits applicability to a narrow set of use cases~\cite{dai2023unsupervised}.


Recently, self-supervised representation learning has emerged as a promising paradigm for learning latent semantic embeddings of speech~\cite{su2023liplearner, rekimoto2023wesper}, mouse movements~\cite{zhang24mouse2vec,zhang24dismouse}, or gaze behaviours~\cite{jiao23_supreyes}.
Self-supervised methods use the original input data as their own supervision without requiring extra human annotations, which are usually costly, cumbersome, and time-consuming~\cite{chu2023wordgesture}.
These methods can also be well generalised across different tasks~\cite{devlin2018bert, he2022masked}.
Despite the advantages and potential of self-supervised representation learning, no prior work has explored such approaches for learning joint representations of human hand and head behaviours.


To fill this gap, we introduce \methodName~-- the first self-supervised method to learn generalisable joint representations of human
hand and head behaviours in extended reality.
At the core of our method is an autoencoder (AE) that uses a graph convolutional network-based (GCN-based) semantic encoder to learn the joint semantic representation of hand-head movements and employs a diffusion-based stochastic encoder to encode the remaining stochastic variations.
Then, a diffusion-based decoder is applied to reconstruct the original signals from the semantic and stochastic representations.
In the training process, we propose to use hand-head movement forecasting as an auxiliary task to enhance the spatial-temporal features encoded in the semantic representation.
We conduct extensive experiments on three publicly available XR datasets, i.e. the EgoBody~\cite{zhang2022egobody}, Aria digital twin (ADT)~\cite{pan2023aria}, and GIMO~\cite{zheng2022gimo} datasets that contain human hand and head data collected during various daily activities from different users in diverse XR environments, and demonstrate that our method significantly outperforms the self-supervised methods that are commonly used for different signals by up to 74.1\% in terms of reconstruction quality and is generalisable across users, activities, and XR environments.
We further show that the semantic representation learned by our method can be used to identify hand-head behaviour clusters with human-interpretable semantics. In contrast, the stochastic representation can generate variable hand-head movements. 
We also evaluate the practical use of our method as a generic feature extractor for two practical sample downstream tasks, i.e. user identification and activity recognition, which are essential for intelligent XR systems to understand users and the interaction context~\cite{hu2022ehtask, hu2021fixationnet, nair2024berkeley, hadnett2019effect}, and demonstrate consistent improvements on both tasks.
The full source code and trained models are available at \url{https://git.hcics.simtech.uni-stuttgart.de/public-projects/HaHeAE}.


\vspace{1em}
\noindent
The specific contributions of our work are three-fold:
\begin{enumerate}[leftmargin=12pt]

\item We present \methodName~ -- a novel self-supervised method that first uses a GCN-based semantic encoder and a diffusion-based stochastic encoder to learn the joint semantic and stochastic embeddings from hand-head signals, respectively, and then applies a diffusion-based decoder to reconstruct the original signals from the learned embeddings. Hand-head forecasting is proposed as an auxiliary training task for refining the semantic representation.

\item We report extensive experiments on three public XR datasets and demonstrate that our method significantly outperforms other methods in reconstruction quality and is generalisable across users, activities, and XR environments.

\item We show that our method enables new applications, including interpretable hand-head cluster identification and hand-head movement generation, and can serve as an effective feature extractor for two practical sample downstream tasks.
\end{enumerate}

%% file: sections/related_work.tex
\section{Related Work}

\subsection{Hand and Head Behaviour Modelling}

Understanding and modelling human behaviours is an essential research topic in extended reality~\cite{hu24_pose2gaze, sidenmark2019eye, jiao23_supreyes, hu2019sgaze}.
Human hand and head movements are particularly significant as they are the most pervasive input modalities in extended reality and are considered crucial components for future human-aware intelligent XR systems~\cite{hu2022ehtask, nair2024berkeley, lin2022Intentional}.
Against this background, researchers have devoted tremendous efforts to the computational modelling of human hand and head behaviours for a wide range of XR applications.
Specifically, Belardinelli et al. modelled human hand behaviour using Gaussian hidden Markov models to predict users' target objects in interactive virtual environments~\cite{belardinelli2022intention}.
Bachynskyi et al. explored second- and third-order lag models for modelling human hand dynamics for mid-air pointing tasks in extended reality~\cite{bachynskyi2020dynamics}.
Gamage et al. proposed a hybrid classical-regressive model to learn hand kinematics for predicting continuous 3D hand trajectory for ballistic movements in immersive virtual environments~\cite{gamage2021so}.
Yang et al. employed a hierarchical Bayesian long short-term memory (LSTM) network to encode historical head movements to predict future head trajectory on omnidirectional images~\cite{yang2021hierarchical}.
Hu et al. used a 1D convolutional neural network (CNN) to extract features from users' head orientations to further predict human visual attention in virtual environments~\cite{hu2020dgaze, hu2021fixationnet, hu25hoigaze}.
Hu et al. proposed to use a combination of a 1D CNN, a recurrent neural network (RNN), and a multi-layer perceptron (MLP) to model human head behaviours for the recognition of user activities in 360-degree videos~\cite{hu2022ehtask}.
However, existing methods for modelling human hand and head behaviours typically explored only a single modality and focused on specific applications rather than learning generalisable representations.
In stark contrast, we are the first to learn generalisable joint representations of human hand and head movements in extended reality.

\subsection{Self-Supervised Representation Learning}

Self-supervised representation learning is a machine learning paradigm where models are trained using only the original input data as supervision information without requiring additional human annotations.
Self-supervised methods have recently been demonstrated to be highly effective for learning generalisable latent semantic embeddings of various signals including speech~\cite{rekimoto2023wesper,su2023liplearner}, 
mouse movements~\cite{zhang24mouse2vec, zhang24dismouse}, and gaze behaviours~\cite{jiao23_supreyes}.
Specifically, Rekimoto et al. used a speech-to-unit encoder to generate hidden speech units and employed a unit-to-speech decoder to reconstruct speech from the encoded speech units~\cite{rekimoto2023wesper}.
Zhang et al. used a Transformer-based encoder-decoder architecture to learn semantic representations of mouse behaviours from both continuous mouse cursor locations and discrete mouse events (click vs. movements)~\cite{zhang24mouse2vec}.
Jiao et al. proposed an implicit neural representation learning-based method to learn semantic representations from low-resolution eye gaze data~\cite{jiao23_supreyes}.
Despite the potential of self-supervised representation learning, no prior work has explored such approaches for learning joint semantic representations of human hand and head behaviours.

\subsection{Denoising Diffusion Models}

Denoising diffusion models, or more precisely, denoising diffusion probabilistic models (DDPM)~\cite{jiao2024diffgaze, jiao2024diffeyesyn} are a class of latent variable generative models consisting of a diffusion process that incrementally adds Gaussian noises to the original input signals and a reverse process that progressively denoises the noisy samples.
At the training stage, a diffusion process is first applied to obtain noisy samples from the original input and then a noise prediction network is trained to predict the noises added to the samples.
At the inference stage, Gaussian samples are used as the input and a reverse process is applied to progressively generate realistic samples from the input by first predicting the noise using the noise prediction network and then denoising the samples.
DDPM models can achieve superior performance over prior methods in terms of generating realistic samples and have been applied in various domains including image and video generation~\cite{ho2022video}, 
time series forecasting~\cite{rasul2021autoregressive}, and human motion prediction~\cite{yan24gazemodiff}.
However, the reverse process of DDPM is a Markov process that progressively samples from an estimated Gaussian distribution to denoise, which is stochastic and slow.
Denoising diffusion implicit model (DDIM)~\cite{song2020denoising} is a variant of DDPM that differs only at the reverse process: the variances of the estimated Gaussian distributions are set to $0$ so that the reverse process becomes deterministic and fast.
By running the reverse process of DDIM backward, an input signal can be deterministically encoded into a noisy sample, which can be seen as a stochastic representation of the original signal that allows high-quality reconstruction.
Owing to its ability for encoding stochastic variations, DDIM has recently been used to learn latent embeddings of various signals~\cite{zhang24dismouse, preechakul2022diffusion}.
For example, Preechakul et al. used an image encoder and a conditional DDIM model to learn the semantic and stochastic embeddings of images respectively and then applied the DDIM model as a decoder to reconstruct the original image from the learned embeddings~\cite{preechakul2022diffusion}.
Despite its potential for representation learning, DDIM has not yet been applied to learn representations of human hand and head behaviours.
In this work, we combine a graph convolutional network and a conditional DDIM to learn the joint semantic and stochastic representations of hand-head movements and use the DDIM model as a decoder to reconstruct the original data.

%% file: sections/method.tex
\section{Method}

\subsection{Problem Definition}

We formulate the problem of learning joint representations of hand and head movements as a self-supervised task.
This task involves learning latent embeddings from the original hand-head movements and using the learned representation to reconstruct the original signals.
Given that different XR environments may use different coordinate systems, we propose to represent hand-head movements using relative coordinates with the origin set to the position of the head.
This approach yields a generalisable representation of hand-head data collected across different XR environments.
We use 3D positions of the left and right hands $ha\in R^{6}$ to represent hand movement following prior works on hand behaviour modelling~\cite{belardinelli2022intention, bachynskyi2020dynamics, gamage2021so} and employ head orientation $he\in R^3$ to denote head movement, where $he$ is a unit vector indicating head forward direction.
Given a sequence of hand movements $HA_{1:N} = \{ha_1, ha_2, ..., ha_N\} \in R^{6\times N}$ and head orientations $HE_{1:N} = \{he_1, he_2, ..., he_N\} \in R^{3\times N}$, the task is to learn latent embeddings that can reconstruct the original signals.

\begin{figure*}
\centering
    \includegraphics[width=1.0\textwidth]{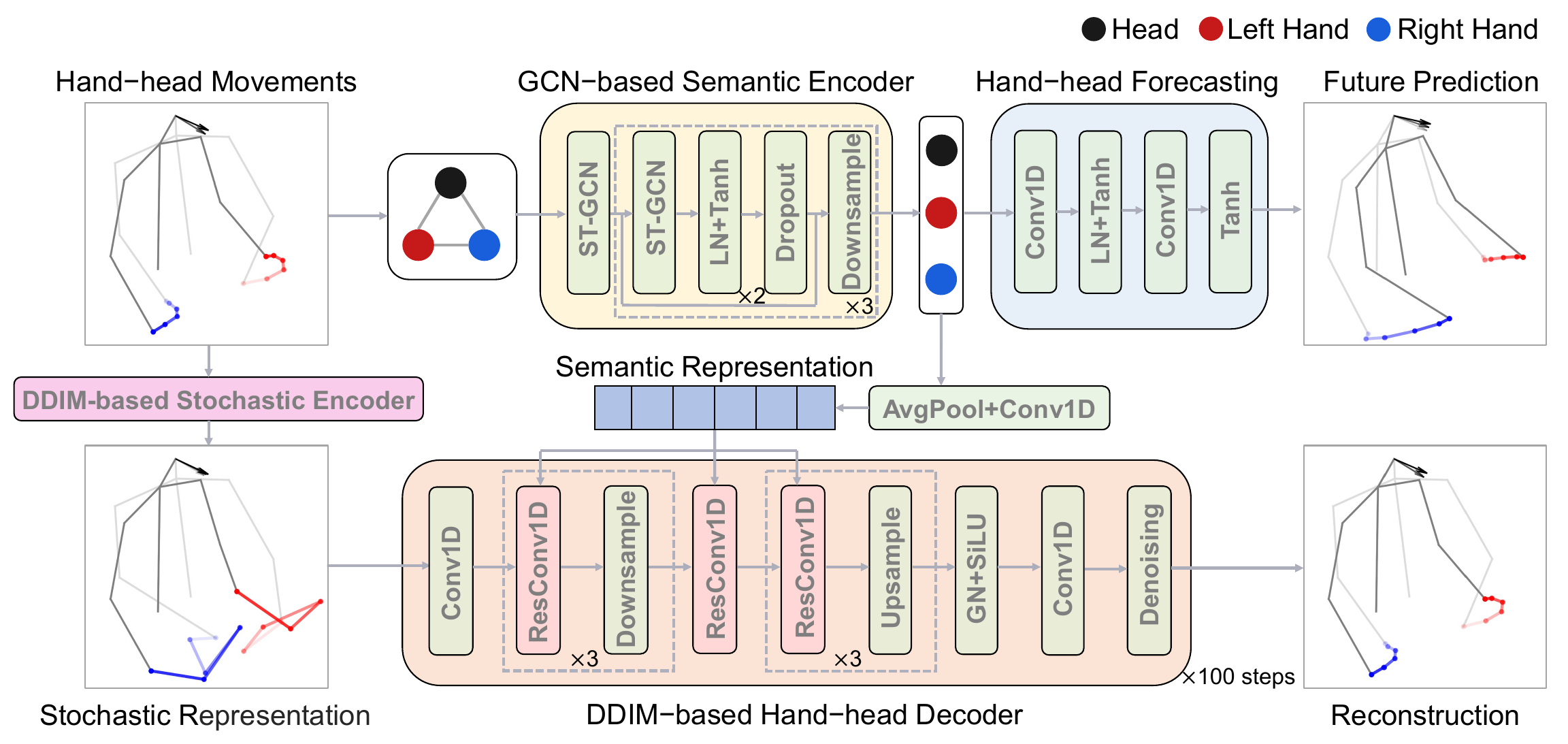}
    \caption{Architecture of \methodName. Our method uses a GCN-based semantic encoder to learn the joint semantic representation of hand-head movements and a DDIM-based stochastic encoder to encode the remaining stochastic variations. A DDIM-based hand-head decoder reconstructs the original input signals from these semantic and stochastic representations. Hand-head forecasting is used as an auxiliary training task to refine the semantic representation.
    }
    \label{fig:method}
\end{figure*}

At the core of our method is an autoencoder that first uses a graph convolutional network-based semantic encoder and a diffusion-based stochastic encoder to learn the joint semantic and stochastic representations of hand-head movements, respectively, and then applies a diffusion-based decoder to reconstruct the original signals.
We propose hand-head forecasting as an auxiliary training task to refine the semantic representation.
See \autoref{fig:method} for an overview of our method.

\subsection{GCN-based Semantic Encoder} \label{sec:semantic_encoder}
Inspired by the fact that graph convolutional networks surpass other architectures such as CNNs or RNNs in learning correlations of different body parts~\cite{ma2022progressively, hu24gazemotion, hu24hoimotion}, we proposed to use a GCN-based encoder to extract semantic features from the hand-head data.
Specifically, we modelled the hand-head data $H_{1:N} = \{ha_1, he_1, ha_2, he_2, ..., ha_N, he_N\} \in R^{9\times N}$ as fully connected spatial and temporal graphs $H \in R^{3\times 3\times N}$ with their adjacency matrices measuring the weights between each pair of nodes.
The spatial graph consists of three joints representing the head, left hand, and right hand, respectively, while the temporal graph contains $N$ nodes corresponding to hand-head data at different time steps.

We first used a spatial-temporal graph convolutional network (ST-GCN) to map the original hand-head data into a latent feature space (see \autoref{fig:method}).
The ST-GCN first performed temporal convolution by multiplying the data with a temporal adjacency matrix $A^T \in R^{N\times N}$, then mapped the original node features ($3$ dimensions) into latent space ($16$ dimensions) using a feature matrix $W\in R^{3\times16}$, and finally performed spatial convolution by multiplying the data with a spatial adjacency matrix $A^S \in R^{3\times 3}$ to obtain hand-head features $H \in R^{16\times 3\times N}$.
We then used a residual GCN module containing two GCN blocks to further process the hand-head features.
Each GCN block consists of an ST-GCN to extract features, a layer normalisation (LN) to normalise the data, a Tanh activation function, and a dropout layer with a dropout rate of $0.1$ to prevent the GCN from overfitting.
The feature matrix of the ST-GCN used in the GCN block was set to $W\in R^{16\times16}$ to ensure that the input and output of the block had the same size.
A residual connection was applied for each GCN block to improve the network flow.
We used three residual GCN modules in total to enhance the hand-head features with a downsample layer applied after the first two modules.
The downsample layer used a 1D average pooling with a kernel size of two to compress the hand-head data along the temporal dimension.
After the downsample operation, we obtained hand-head features in the size of $H \in R^{16\times 3\times (N/4)}$.

We finally aggregated the hand-head features along the spatial dimension ($R^{16\times 3\times (N/4)}\rightarrow R^{48\times (N/4)}$), applied an adaptive average pooling to compress the features into a sequence length of one ($R^{48\times (N/4)}\rightarrow R^{48\times 1}$), and used a 1D convolution layer with $128$ channels to map the compressed features into a semantic feature vector $E_{sem}\in R^{128}$.

\subsection{DDIM-based Hand-head Decoder}\label{sec:decoder}

We used a conditional DDIM as the decoder to reconstruct the original hand-head movements from the input of $(H_T, T, E_{sem})$, where $H_T$ is the stochastic representation of the original hand-head data, $T$ is the number of denoising steps and is set to $100$ following common settings~\cite{preechakul2022diffusion, zhang24dismouse, yan24gazemodiff}, and $E_{sem}$ is the semantic representation.
At the core of our decoder are a noise prediction network $\epsilon_\theta$ that estimates the noise added to the original data and a denoising process that iteratively generates a cleaner version based on the estimated noise.

\begin{figure}
\centering
    \includegraphics[width=0.48\textwidth]{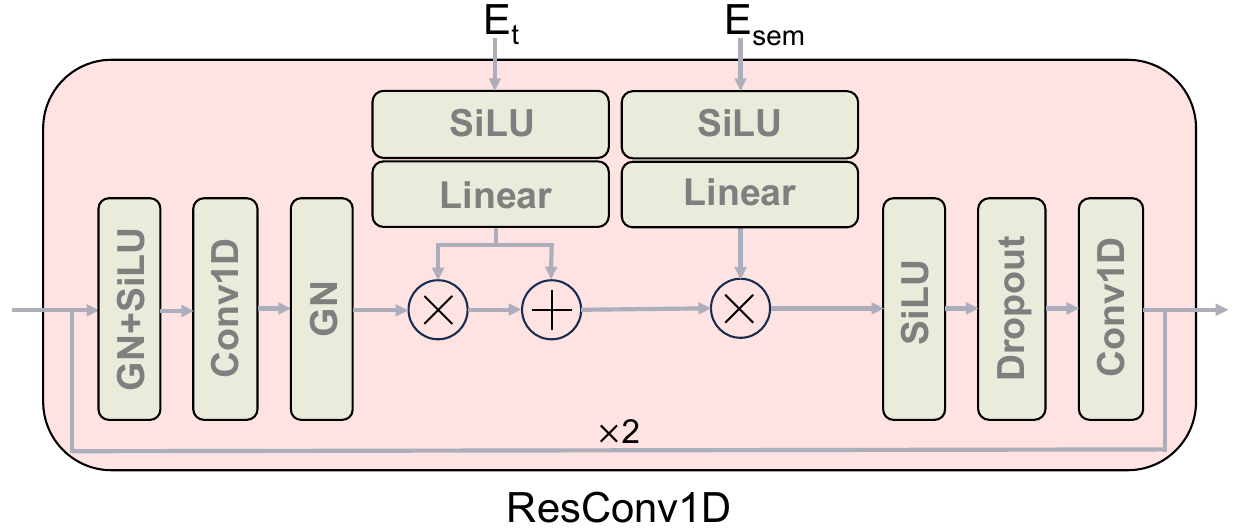}
    \caption{The ResConv1D module used in our method is conditioned on the time step embedding $E_t$ and the semantic representation $E_{sem}$.}
    \label{fig:resconv1d}
\end{figure}

\paragraph{Noise Prediction Network}
We used a 1D CNN-based UNet as our noise prediction network (see \autoref{fig:method}) in light of the good performance of UNet as a backbone in diffusion models~\cite{preechakul2022diffusion, dhariwal2021diffusion}.
We employed 1D CNN because it is computationally efficient and performs well for processing time series data~\cite{hu2020dgaze, hu2021fixationnet, hu2022ehtask}.
Specifically, we first used a 1D CNN layer with $64$ channels and a kernel size of three to convert the input signal $H_t \in R^{9\times N}$ into a latent feature space.
We then converted the time step $t$ to an embedding $E_t$ by sequentially applying a sinusoidal encoding, a linear layer that has $128$ neurons, a SiLU activation function, and a linear layer with $128$ neurons.
We further employed a ResConv1D module to extract features by using the time step embedding $E_t$ and the semantic representation $E_{sem}$ as the condition, as illustrated in \autoref{fig:resconv1d}.
The ResConv1D module contains two residual blocks, each consisting of a group normalisation (GN)~\cite{wu2018group}, a SiLU activation function, a 1D CNN layer with $64$ channels and a kernel size of three, a group normalisation, a SiLU function, a dropout layer with a dropout rate of $0.1$, and a 1D CNN layer with $64$ channels and a kernel size of three.
$E_{t}$ was processed by a SiLU function and a linear layer to obtain an embedding in the size of $128$.
Half of $E_{t}$ (in the size of $64$) was multiplied with the output of the second GN layer while the other half was added to the features, following prior works that used time step as a condition~\cite{dhariwal2021diffusion, preechakul2022diffusion}.
$E_{sem}$ was processed by a SiLU function and a linear layer with $64$ neurons and was then applied after $E_t$ as the second condition.
In addition to the ResConv1D module, we also applied downsample layers to compress the data along the temporal dimension and symmetrically used upsample layers to recover the data following common practice of UNet~\cite{zhang24dismouse, preechakul2022diffusion, dhariwal2021diffusion}.
We finally used a GN layer, a SiLU function, and a 1D CNN layer with nine channels and a kernel size of three to estimate the noise.

\paragraph{Denoising Process}
Following prior settings in DDIMs~\cite{song2020denoising, preechakul2022diffusion}, our decoder used the following deterministic denoising process:
\begin{equation}
\label{eq:ddim_denoise}
    H_{t-1}=\sqrt{\alpha_{t-1}}(\frac{H_t-\sqrt{1-\alpha_t}\epsilon_\theta^t}{\sqrt{\alpha_t}}) + \sqrt{1-\alpha_{t-1}}\epsilon_\theta^t,
\end{equation}
where $\epsilon_\theta^t = \epsilon_\theta(H_t, t, E_{sem})$ is the estimated noise, $\alpha_t = \prod_{i=1}^{t}(1-\beta _i)$ and $\beta_1, ..., \beta_t \in (0, 1)$ are hyper-parameters used to control the noise level at each step.
By running the denoising process iteratively for $T$ steps, the original signals can be reconstructed from the stochastic and semantic hand-head representations.

\subsection{DDIM-based Stochastic Encoder}\label{sec:stochastic_encoder}

After being trained, our conditional DDIM described in \autoref{sec:decoder} can also be used as an encoder~\cite{preechakul2022diffusion, zhang24dismouse} to obtain the stochastic representation $E_{sto} = H_T$ from the original hand-head data $H_0$ by running its deterministic denoising process backward (the reverse of \autoref{eq:ddim_denoise}) for $T$ steps:
\begin{equation}
\label{eq:ddim_encode}
    H_{t+1}=\sqrt{\alpha_{t+1}}(\frac{H_t-\sqrt{1-\alpha_t}\epsilon_\theta^t}{\sqrt{\alpha_t}}) + \sqrt{1-\alpha_{t+1}}\epsilon_\theta^t.
\end{equation}
$E_{sto}$ and $E_{sem}$ complement each other by respectively encoding the stochastic details and high-level semantics of the original hand-head signals.
Note that $E_{sto}$ is only used in the inference stage, i.e. after the DDIM has been trained, to allow high-quality reconstruction of the hand-head movements.

\subsection{Hand-head Forecasting} \label{sec:hand_head_forecasting}

Prior works on large language models (LLMs) have demonstrated the effectiveness of next token prediction as a training task to learn semantic representations of sequential data~\cite{touvron2023llama, brown2020language}.
Inspired by this, we proposed to use hand-head forecasting as an auxiliary training task to enhance the spatial-temporal features encoded in the semantic representation learned by our method, as illustrated in \autoref{fig:method}.
In light of the good performance of 1D CNN for human behaviour forecasting~\cite{hu2020dgaze,hu2021fixationnet}, we used the features learned by our semantic encoder as input to 1D CNNs to predict hand-head movements in the near future $H_{N+1:N+\Delta n} = \{ha_{N+1}, he_{N+1}, ha_{N+2}, he_{N+2}, ..., ha_{N+\Delta n}, he_{N+\Delta n}\} \in R^{9\times \Delta n}$.
Specifically, for hand movement forecasting we applied two 1D CNN layers having $32$ channels with kernel size of three and six channels with kernel size of one respectively to process the hand features while for head movement forecasting we used two 1D CNN layers with $16$ channels, whose kernel size is three, and three channels, whose kernel size is one, respectively to process the head features.
A layer normalisation and a Tanh activation function were applied after the first CNN layer while a Tanh activation function was employed after the second CNN layer.
The predicted head movements were normalised to unit vectors to represent head orientations.

\subsection{Training and Inference}
\paragraph{Training Stage}
At the training stage, we first generated a noisy version $H_t$ of the original hand-head data $H_0$ by adding random noise to it using the following formula:
\begin{equation}
\label{eq:ddim_forward}
    H_{t}=\sqrt{\alpha_{t}}H_0 + \sqrt{1-\alpha_{t}}\epsilon,
\end{equation}
where $\epsilon$ is a random noise sampled from $N(0, I)$ and has the same size as $H_0$, $t$ is the number of time steps and is uniformly sampled from \numrange{1}{1000}.
We further used $(H_t, t, E_{sem})$ as input to the noise prediction network $\epsilon_\theta$ to estimate the noise.
We finally trained our method in an end-to-end manner using the combination of a noise prediction loss $L_{noise}$ and a hand-head forecasting loss $L_{forecasting}$:
\begin{align*}
L &= L_{noise}+L_{forecasting}\\
&=\frac{1}{N} \left \|  \epsilon_\theta(H_t, t, E_{sem})-\epsilon  \right \|^2+\frac{1}{\Delta n} \left \| \hat{H}_{future} - H_{future} \right \|^2,
\end{align*}
where 
$\hat{H}_{future} = \hat{H}_{N+1:N+\Delta n}$ and $H_{future} = H_{N+1:N+\Delta n}$ are the predicted and ground truth future movements respectively.

\paragraph{Inference Stage}
At the inference stage, we first calculated the semantic representation $E_{sem}$ from the original hand-head data $H_0$, then used $(H_0, T, E_{sem})$ as input to the stochastic encoder to obtain the stochastic representation $E_{sto} = H_T$, and finally reconstructed the original signal from $(H_T, T, E_{sem})$ using the hand-head decoder.

%% file: sections/experiments.tex
\section{Experiments and Results} \label{sec:experiments}


\subsection{Datasets} \label{sec:datasets}

We evaluated our method on three publicly available XR datasets, i.e. the EgoBody~\cite{zhang2022egobody}, ADT~\cite{pan2023aria}, and GIMO~\cite{zheng2022gimo} datasets that contain human hand and head data collected during various daily activities from different users in diverse XR environments.

\paragraph{EgoBody}
The EgoBody dataset collects the hand and head movements from $36$ users performing various social interaction activities, including \textit{catch}, \textit{chat}, \textit{dance}, \textit{discuss}, \textit{learn}, and \textit{teach}, in $15$ indoor environments.
The data was recorded using a Microsoft HoloLens2 headset at $30$ fps.
The dataset contains $125$ sequences, each lasting about 2 minutes.

\paragraph{ADT}
The ADT dataset contains $34$ sequences of human hand and head movements performing three indoor activities, i.e. \textit{decoration}, \textit{meal}, and \textit{work}, in two virtual environments, including an apartment and an office environment.
The data was collected using an Aria glass at $30$ fps and each sequence lasts around 2 minutes.

\paragraph{GIMO}
The GIMO dataset records hand and head data from $11$ users performing various daily activities in $19$ indoor scenarios.
The activities cover three categories, i.e. \textit{change the state of objects} (\textit{open}, \textit{push}, \textit{transfer}, \textit{throw}, \textit{pick up}, \textit{lift}, \textit{connect}, \textit{screw}, \textit{grab}, \textit{swap objects}), \textit{interact with objects} (\textit{touch}, \textit{hold}, \textit{step on}, \textit{reach to objects}), and \textit{rest} (\textit{sit} or \textit{lay on objects}).
The data was collected using a Microsoft HoloLens2 headset at $30$ fps.
The whole dataset contains $215$ sequences, each lasting about 10 seconds.

\paragraph{Training and Test Sets}
We trained our method using the EgoBody dataset since it is the largest among the three datasets.
Specifically, we followed the default training and test splits in the original paper~\cite{zhang2022egobody}, i.e. using $82$ sequences for training and the remaining $43$ sequences for testing.
The training and test sets of EgoBody have no overlapping users.
We also used two unseen datasets, ADT and GIMO, to evaluate our method's generalisation capability for different users, activities, and XR environments.

\subsection{Evaluation Settings} \label{sec:eval_settings}

\paragraph{Evaluation Metrics}
For reconstructing hand trajectories, we followed prior works on human motion modelling~\cite{hu24gazemotion, hu24hoimotion} to use the mean per joint position error (MPJPE), i.e. the mean of the left and right-hand position errors, as the evaluation metric.
For reconstructing head orientations, we used the mean angular error between the reconstructed and ground truth head orientations as our metric following prior works on head behaviour modelling~\cite{hu2019sgaze,hu2020dgaze}.

\paragraph{Baselines}
To the best of our knowledge, no methods exist for learning semantic representations of human hand and head movements.
To better evaluate our method, we used variational autoencoders (VAEs), which are commonly used for learning semantic representations~\cite{vahdat2020nvae, leglaive2020recurrent, li2023causal}, as the baselines to compare with our method.
VAE consists of an encoder that learns the semantic embeddings from the input data and a decoder to reconstruct the original signals.
The encoder and decoder of a VAE can be implemented using arbitrary neural networks.
To provide a comprehensive comparison, we implemented variational autoencoders using four commonly used neural networks:

\begin{itemize}[noitemsep,leftmargin=*]    
    \item VAE\_1DCNN: The encoder consists of three 1D CNN layers, followed by a layer normalisation and a ReLU activation function. Each CNN layer has $32$ channels and a kernel size of three.
    The decoder contains three 1D transposed CNN layers: the first two transposed CNN layers have $32$ channels and a kernel size of three, each followed by a layer normalisation and a ReLU activation function. In contrast, the third transposed CNN layer has a kernel size of three and a channel size equal to the input dimension to reconstruct the original data.
    
    \item VAE\_LSTM: The encoder contains an LSTM layer with $32$ channels, while the decoder has an LSTM layer with a channel size equal to the input dimension for reconstruction.
    
    \item VAE\_GRU: The encoder has a GRU layer with $32$ channels, while the decoder uses a GRU layer with a channel size equal to the input dimension for reconstructing the data.
    
    \item VAE\_MLP: The encoder consists of two linear layers with $128$ neurons, each followed by a layer normalisation and a ReLU activation function.
    The decoder contains two linear layers: the first linear layer has $128$ neurons. A layer normalisation and a ReLU activation function follow it. In contrast, the second linear layer has an output size equal to the input dimension to reconstruct the original signals.
\end{itemize}

\paragraph{Implementation Details}
We set the sequence length of hand-head movements $N$ to $40$ and the time horizon of hand-head forecasting $\Delta n$ to $3$.
We trained our method and the baseline methods for a total of $130$ epochs using the Adam optimiser with a learning rate of $1e-4$ and a batch size of $64$.
Our method was implemented on an NVIDIA Tesla V100 SXM2 32GB GPU with an Intel(R) Xeon(R) Platinum 8260 CPU @ 2.40GHz and it took around nine hours for our method to finish its training on the EgoBody dataset.

\subsection{Reconstruction Results}

\paragraph{Results on EgoBody}
We first trained and tested our method on the EgoBody dataset and indicated the reconstruction performances of different methods in \autoref{tab:results}.
The table shows the MPJPE error (in centimetres) of hand trajectory reconstruction and the mean angular error (in degrees) of head orientation reconstruction.
As can be seen from the table, our method significantly outperforms other methods on EgoBody, achieving an improvement of $51.8\%$ ($1.664$ vs. $3.455$) and $64.3\%$ ($0.834$ vs. $2.338$) in terms of hand and head reconstruction, respectively.
We further performed a paired Wilcoxon signed-rank test by first generating a distribution of performance metrics for each method and then comparing them directly.
The test sample size of the EgoBody dataset is $66293$.
The results validated that the differences between our method and the baselines are statistically significant ($p<0.01$).
We also analysed the distributions of different methods' reconstruction errors and validated that our method achieves better performance than other methods (see supplementary material for more details).

\begin{table}[t]
	\centering
	\caption{Hand and head reconstruction errors (hand unit: centimetres, head unit: degrees) of different representation methods on the EgoBody, ADT, and GIMO datasets. Best results are in bold.
    } \label{tab:results}
        \resizebox{0.45\textwidth}{!}{
	\begin{tabular}{ccccccc}
		\toprule
    &\multicolumn{2}{c}{\textbf{EgoBody}} &\multicolumn{2}{c}{\textbf{ADT}} &\multicolumn{2}{c}{\textbf{GIMO}} \\ \cmidrule(lr){2-3} \cmidrule(lr){4-5} \cmidrule(lr){6-7}
    & hand & head & hand & head & hand & head\\ \midrule
  VAE\_1DCNN &3.575 &2.549 &3.876 &2.776 &4.422 &3.100 \\ 
  VAE\_LSTM &7.254 &5.421 &7.933 &9.928 &8.908 &8.060 \\ 
  VAE\_GRU &6.776 &4.390 &7.351 &6.161 &8.369 &6.371 \\ 
  VAE\_MLP &3.455 &2.338 &3.932 &2.733 &
  4.310 &2.927 \\ \midrule  
  Ours &\textbf{1.664} &\textbf{0.834} &1.966 &\textbf{0.707} &\textbf{2.397} &\textbf{1.247} \\
  Ours\_1DCNN &2.010 &1.070 &2.370 &1.095 &2.883 &1.500 \\
  Ours\_LSTM &1.706 &0.842 &\textbf{1.937} &0.713 &2.587 &1.341\\
  Ours\_GRU &1.715 &0.861 &1.964 &0.718 &2.658 &1.377\\
  Ours\_MLP &1.840 &0.851 &2.213 &0.822 &2.660 &1.279 \\  
  Ours w/o $E_{sem}$ &56.803 &93.361 &55.952 &92.991 &57.005 &91.864\\
  Ours w/o $E_{sto}$ &10.604 &11.278 &11.889 &12.878 &11.158 &12.042\\      
   \bottomrule
    \end{tabular}}
\end{table}

\paragraph{Results on ADT and GIMO}
We further tested our method and the baseline methods, which were trained on EgoBody, directly on the ADT and GIMO datasets without any fine-tuning.
The hand and head reconstruction performances of different methods are summarised in \autoref{tab:results}.
We can see from the table that our method consistently outperforms other methods by a large margin on both the ADT and GIMO datasets.
Specifically, on the ADT dataset, our method outperforms other methods by $49.3\%$ ($1.966$ vs. $3.876$) and $74.1\%$ ($0.707$ vs. $2.733$) in terms of hand and head reconstruction, respectively.
On the GIMO dataset, our method achieves an improvement of $44.4\%$ ($2.397$ vs. $4.310$) and $57.4\%$ ($1.247$ vs. $2.927$) in hand and head reconstruction, respectively.
A paired Wilcoxon signed-rank test was further performed to compare the performances of our method with other methods.
The test sample sizes of the ADT and GIMO datasets are $91998$ and $44081$, respectively.
The results demonstrated that the differences between our method and other methods on both the ADT and GIMO datasets are statistically significant ($p<0.01$).

\paragraph{The Effectiveness of the GCN-based Semantic Encoder}
To test the effectiveness of our GCN-based encoder, we replaced it with the baseline VAE encoders (1DCNN, LSTM, GRU, or MLP) to re-train our method.
We can see from \autoref{tab:results} that our GCN-based encoder achieves better performance than other encoders in terms of hand-head reconstruction quality, demonstrating its superiority for encoding hand-head signals.

\paragraph{The Effectiveness of the Semantic and Stochastic Representations}
We also evaluated the effectiveness of $E_{sem}$ and $E_{sto}$ in reconstructing the original signals.
Specifically, we first replaced $E_{sem}$ and $E_{sto}$ with Gaussian noise respectively and then used them to reconstruct the hand-head data, following prior works on evaluating representations~\cite{zhang24dismouse, preechakul2022diffusion}.
We can see from \autoref{tab:results} that the reconstruction performance deteriorates significantly after removing $E_{sem}$ or $E_{sto}$, validating that both $E_{sem}$ and $E_{sto}$ are significant for reconstructing the original hand-head data.
We also find that $E_{sem}$ has a notably higher influence on the reconstruction quality than $E_{sto}$.
This is because $E_{sem}$ captures the semantic information that is critical for reconstruction while $E_{sto}$ encodes the remaining stochastic variations that are less important~\cite{preechakul2022diffusion, zhang24dismouse}.



%% file: sections/applications.tex
\section{Use Cases Enabled By Our Method}


\subsection{Identifying Interpretable Hand-head Clusters} \label{sec:application_clusters}

\begin{figure*}[htp]
    \centering
    \begin{subfigure}[b]{0.24\textwidth}
        \centering
        \includegraphics[width=\textwidth]{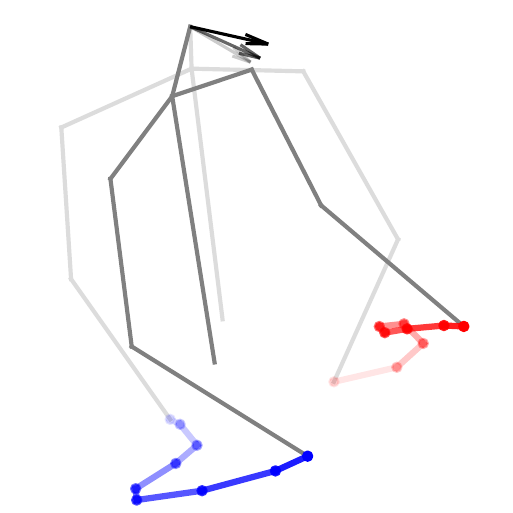}
        \caption{Activity: \textbf{Instruct to act}. The head is facing slightly downward; both hands move noticeably below the head.}%
        \label{fig:cluster1}
    \end{subfigure}
    \begin{subfigure}[b]{0.24\textwidth}
        \centering
        \includegraphics[width=\textwidth]{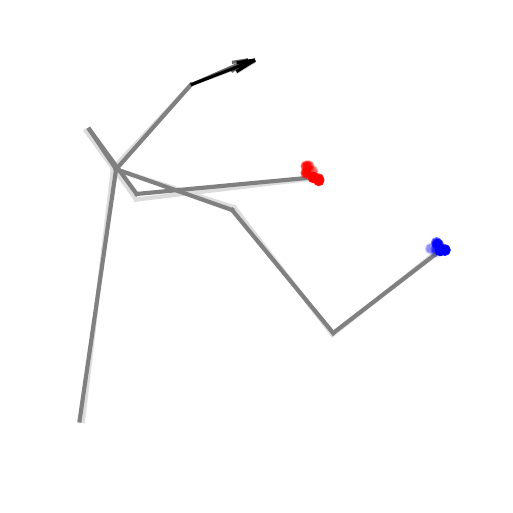}
        \caption{Activity: \textbf{Learn course while sitting}. The head is facing slightly upward; both arms are bent and have no large movements.}
        \label{fig:cluster2}
    \end{subfigure}
    \begin{subfigure}[b]{0.24\textwidth}
        \centering
        \includegraphics[width=\textwidth]{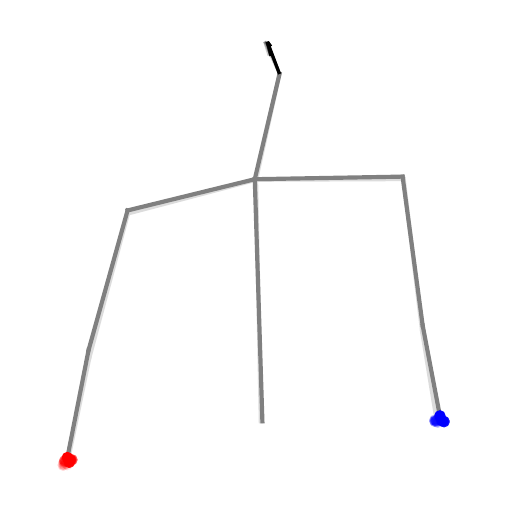}
        \caption{Activity: \textbf{Casually chat while standing}. The head is facing upward; both arms are laid down and remain almost still.}
        \label{fig:cluster3}
    \end{subfigure}
    \begin{subfigure}[b]{0.24\textwidth}
        \centering
        \includegraphics[width=\textwidth]{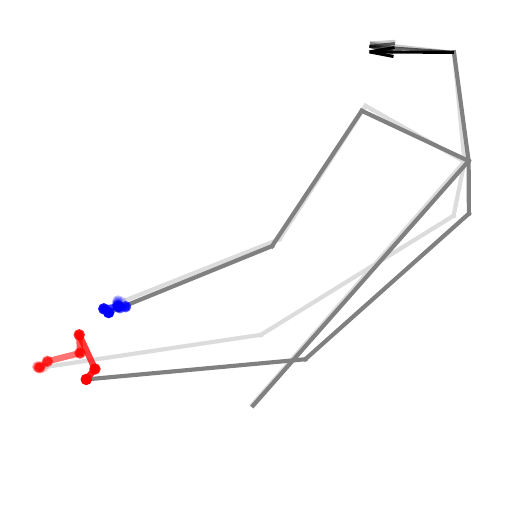}
        \caption{Activity: \textbf{Take a tape}. The head is facing forward; the left hand has a greater range of motion than the right hand.}
        \label{fig:cluster4}        
    \end{subfigure}
    \caption{Representative hand and head movements for each of the four largest clusters and their semantics on the EgoBody dataset.
    The red and blue lines indicate the trajectories of left and right hands, respectively, while the black arrows denote head orientations. The colours of the lines and arrows are gradually deepened over time.
    }
    \label{fig:clusters}
\end{figure*}

Analysing and understanding human hand and head behaviours is a significant research topic in extended reality and is considered a crucial component for future human-aware intelligent XR systems~\cite{hu2021fixationnet, hadnett2019effect, hu2022ehtask, nair2024berkeley}.
This section shows that our method can facilitate hand-head behaviour analysis by identifying hand-head behaviour clusters with human-interpretable semantics.
Specifically, we first calculated the semantic representations of the hand-head movements on the test set of EgoBody using our method and then identified hand-head clusters in the semantic embedding space using hierarchical density-based spatial clustering of applications with noise (HDBSCAN)~\cite{campello2013density}.
HDBSCAN is a clustering method that can handle clusters with varying densities and noisy data points and has shown good performance in representation-based clustering~\cite{yilma2023elements, zhang24mouse2vec}.
We followed common practice in representation learning to use cosine similarity as the distance metric for HDBSCAN to find clusters~\cite{li2022structure,han2020modelling,zhang24mouse2vec}.
In this way, we identified a total of $176$ clusters on the test set of EgoBody.
We examined the four clusters with the largest data samples for simplicity since they provided insights into the most prevalent behaviour patterns.
For each cluster, we used the hand-head data sample that is closest to the cluster's centroid as the representative of the whole cluster~\cite{leisch2006toolbox,zhang24mouse2vec} to further analyse and understand human hand-head behaviours.
The representative hand and head movements for each of the four largest clusters and their semantics are illustrated in \autoref{fig:clusters} with red and blue lines indicating hand trajectories and black arrows denoting head orientations.

We observed that the largest cluster (a) mainly consists of noticeable horizontal movements of both hands across different activities such as \textit{instruct to act} or \textit{perform}.
Furthermore, we found that cluster (b) and cluster (c) contain very limited hand-head movements at different locations, suggesting that the user mainly stood or sat still and barely moved.
The samples forming these clusters are mainly generated during the activities that require few hand and head movements such as \textit{learn course while sitting} or \textit{casually chat while standing}.
In these scenarios, users mainly stay still and may only use slight hand or head movements (e.g., gestures or nodding) as the body language during communications~\cite{duarte2018action}, in line with the patterns illustrated by these two clusters.
In addition, cluster (d) demonstrates moving one hand and the other hand mainly staying still.
These samples were generated from activities requiring only one hand to interact with objects such as \textit{take a tape} or \textit{pick up a cup}.
The above results demonstrate that our method can be applied to identify common patterns of hand-head movements and thus has great potential to facilitate the analysis of hand-head signals.
Furthermore, we also compared the clustering performance of our method with that of the baseline methods and validated the superiority of our method (see supplementary material for more details).

\subsection{Generating Variable Hand-head Movements} \label{sec:application_generation}

\begin{figure*}[htp]
    \centering
    \captionsetup[subfigure]{labelformat=empty}
    \begin{subfigure}[b]{0.19\textwidth}
        \centering
        \includegraphics[width=\textwidth]{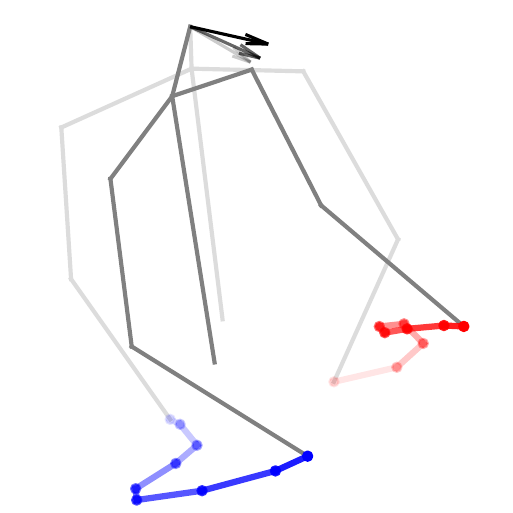}        
        \newline\includegraphics[width=\textwidth]{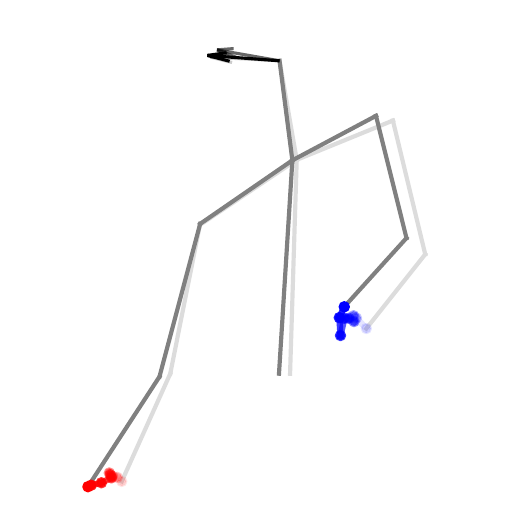}        
        \newline\includegraphics[width=\textwidth]{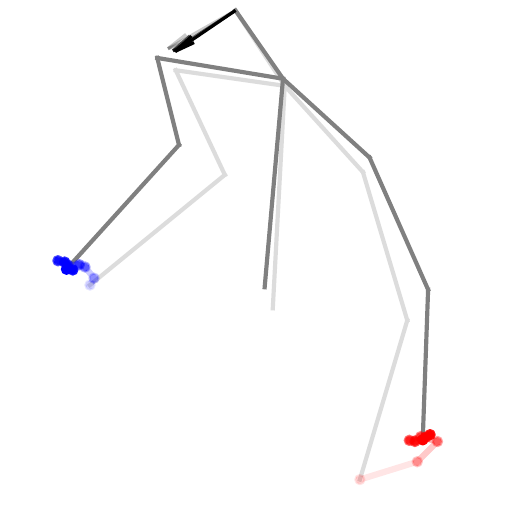}        
        \caption{Original}%
    \end{subfigure}
    \begin{subfigure}[b]{0.19\textwidth}
        \centering
        \includegraphics[width=\textwidth]{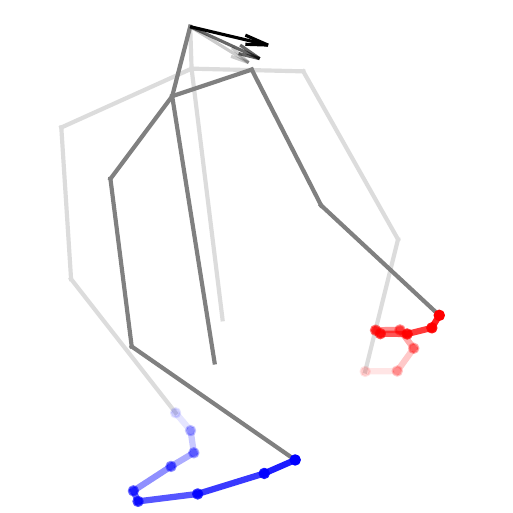}        
        \newline\includegraphics[width=\textwidth]{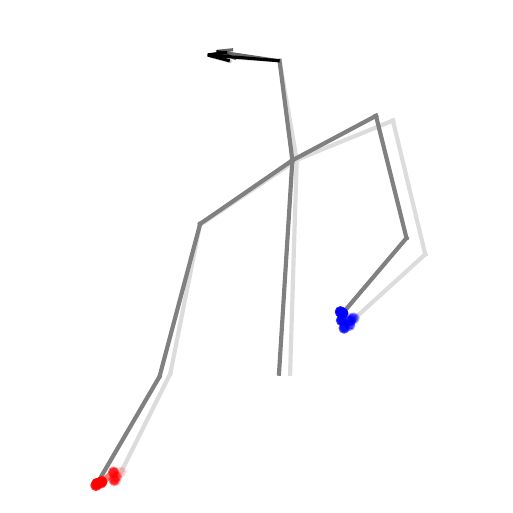}
        \newline\includegraphics[width=\textwidth]{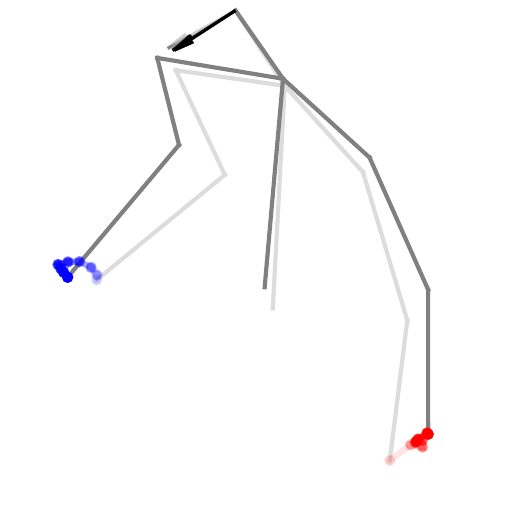}        
        \caption{$E_{sto}$}
    \end{subfigure}
    \begin{subfigure}[b]{0.19\textwidth}
        \centering
        \includegraphics[width=\textwidth]{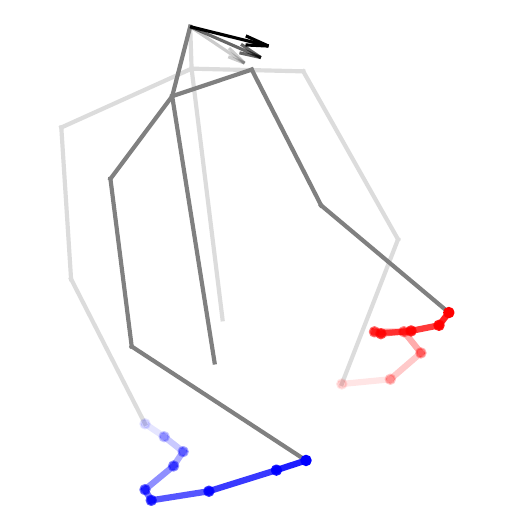}        
        \newline\includegraphics[width=\textwidth]{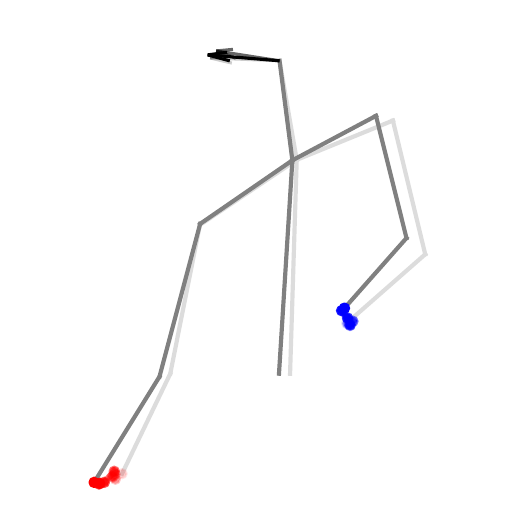}
        \newline\includegraphics[width=\textwidth]{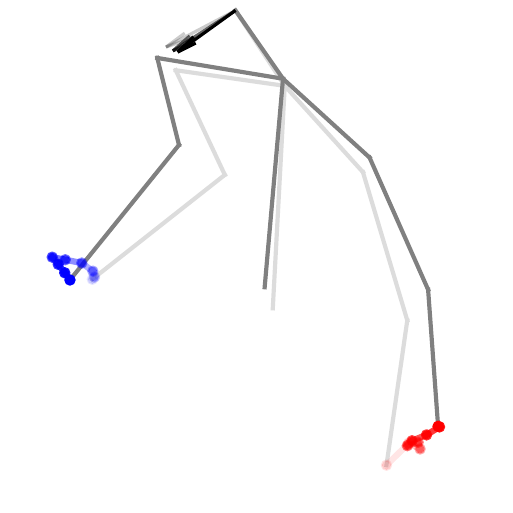}        
        \caption{$E_{sto}$ + $0.1\cdot z$}
    \end{subfigure}
    \begin{subfigure}[b]{0.19\textwidth}
        \centering
        \includegraphics[width=\textwidth]{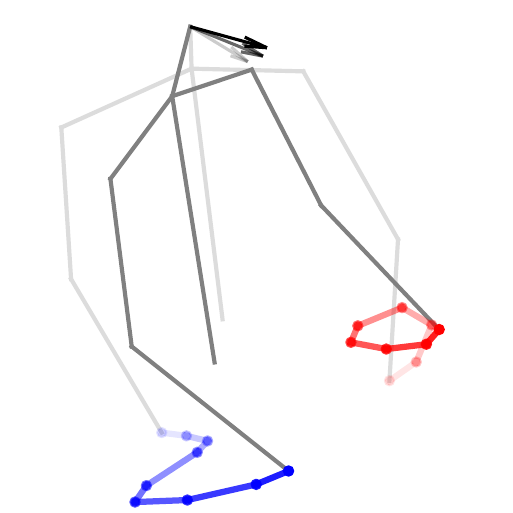}        
        \newline\includegraphics[width=\textwidth]{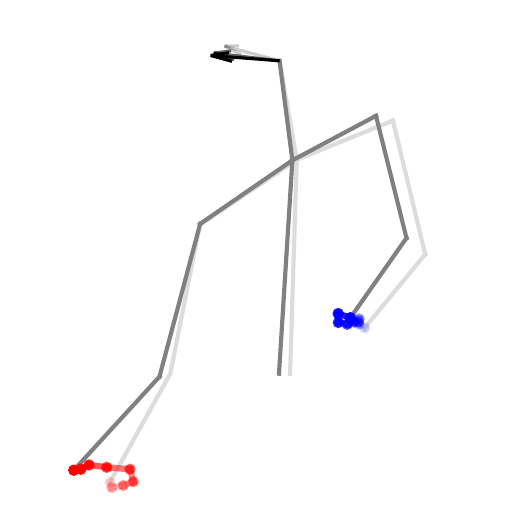}
        \newline\includegraphics[width=\textwidth]{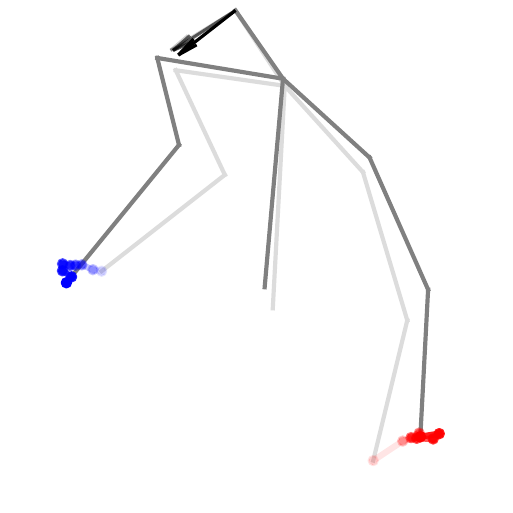}        
        \caption{$E_{sto}$ + $0.3\cdot z$}
    \end{subfigure}    
    \begin{subfigure}[b]{0.19\textwidth}
        \centering
        \includegraphics[width=\textwidth]{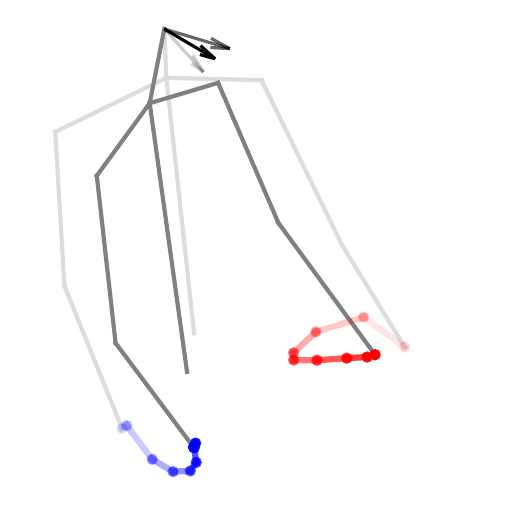}        
        \newline\includegraphics[width=\textwidth]{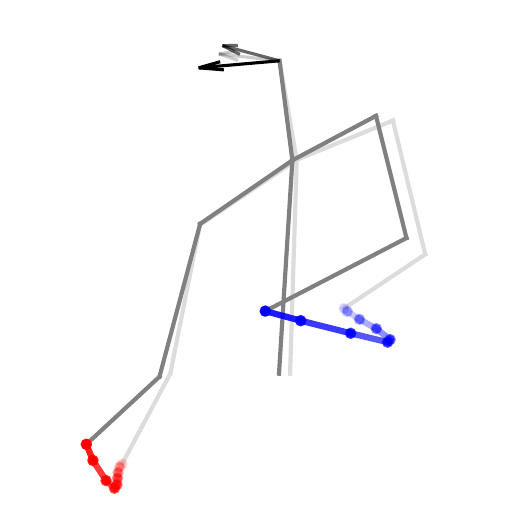}
        \newline\includegraphics[width=\textwidth]{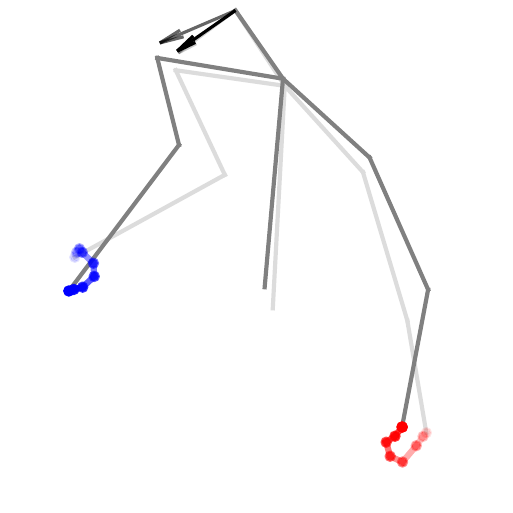}
        \caption{$z$}
    \end{subfigure}
    \caption{Hand and head movements generated from altered stochastic representations and random noise on the EgoBody dataset. Our method can generate variable and realistic hand-head movements from altered stochastic representations and random noise.
    }
    \label{fig:generation}
\end{figure*}

Collecting large-scale human behavioural data in extended reality serves as the basis for understanding human behaviours~\cite{hu2019sgaze,hu2020dgaze,hu2021fixationnet} and developing data-driven models~\cite{nair2024berkeley, hu24_pose2gaze,hu2022ehtask}.
However, human behavioural data collection is usually costly and time-consuming and raises serious concerns about privacy and security~\cite{elfares22_federated, nair2024berkeley, elfares24_privateyes}.
To address these challenges, one common solution is to augment existing datasets instead of collecting new data~\cite{zhang24mouse2vec, zhang24dismouse}.
Our method can be used for data augmentation by generating hand-head movements with controllable randomness from the original signals.
Specifically, we first added random noise to the stochastic representation $E_{sto}$ learned by our method using the formula:
\begin{equation}
\label{eq:ftheta}
    E_{var} = E_{sto} + \beta\cdot z,
\end{equation}
where $z$ is a random noise sampled from $N(0, I)$ and has the same size as $E_{sto}$, \hl{$\beta$ is a fixed constant used as a weighting factor to control the randomness}, and $E_{var}$ is the altered stochastic representation.
We then used our decoder (see \autoref{fig:method}) to generate hand-head movements from the semantic and altered stochastic representations.
This way, our method can generate hand-head movement variations that differ in details while preserving the core semantic information.
\autoref{fig:generation} shows some generation examples on the EgoBody dataset.
We can see that the generated hand-head movements are similar to the original one in overall moving trend, with notable differences in moving trajectory.
To further explore our method’s applicability, we replaced our stochastic representation with random noise $z$ to generate hand-head signals. 
The results in \autoref{fig:generation} (column $z$) showed that our method can generate variable and realistic hand-head movements from random noise.
The above results illustrate the usefulness of our method as an effective means to generate variable and realistic hand-head movements, which has significant potential for enriching and augmenting existing hand-head movement datasets.

\subsection{Serving as a Reusable Feature Extractor} \label{sec:application_downstream}

A pre-trained self-supervised representation method can usually be reused for relevant downstream tasks by using the semantic representation learned by the method as pre-computed features~\cite{zhang24mouse2vec,zhang24dismouse,preechakul2022diffusion,li2021screen2vec}.
Inspired by this, we further evaluated the practical use of our method as a generic feature extractor for two sample downstream tasks, i.e. user identification and activity recognition which are essential for intelligent XR systems to understand users and the interaction context~\cite{hu2022ehtask, nair2024berkeley, hadnett2019effect, hu2021fixationnet}.

\begin{table}[t]
	\centering
	\caption{Performances of the semantic representations learned by different methods on the downstream tasks of user identification and activity recognition. Best results are marked in bold. The representation learned by our method significantly outperforms the representations generated by other methods in terms of both user identification accuracy and activity recognition accuracy. Our method consistently outperforms its ablated versions, validating the effectiveness of each component used in our method.}\label{tab:downstream}
        \resizebox{0.45\textwidth}{!}{
	\begin{tabular}{cccc}
		\toprule
    &\multicolumn{1}{c}{\textbf{User Identification}} &\multicolumn{2}{c}{\textbf{Activity Recognition}}  \\ \cmidrule(lr){2-2} \cmidrule(lr){3-4} 
    &EgoBody &EgoBody &ADT \\ \midrule
  Chance &8.3\% &33.3\% &33.3\% \\ \midrule
  VAE\_1DCNN &26.3\% &48.8\% &62.7\% \\ 
  VAE\_LSTM &24.9\% &47.1\%  &61.3\% \\ 
  VAE\_GRU &28.0\% &41.3\% &61.0\% \\ 
  VAE\_MLP &25.8\% &50.5\% &60.7\% \\ \midrule 
  Ours \textit{hand only} &18.0\% &55.5\% &53.6\%\\
  Ours \textit{head only} &25.7\% &46.1\% &62.5\%\\  
  Ours w/o \textit{forecasting} &29.4\% &54.7\% &63.1\% \\
  Ours &\textbf{29.8\%} &\textbf{55.7\%} &\textbf{63.9\%} 
  \\ \bottomrule
    \end{tabular}}
\end{table}

\subsubsection{User Identification} \label{sec:user_identification}

User identification is a popular research topic in extended reality and has great relevance for a variety of applications such as adapting virtual environments~\cite{liebers2021understanding}, personalising user interfaces~\cite{liebers2021understanding}, and authenticating users~\cite{liebers2021understanding, nair2023unique}.
Recent studies have revealed that a user's hand and head movements in extended reality can be used as a means to identify the user~\cite{liebers2021understanding, nair2023unique}.
Therefore, user identification can be used as a relevant sample downstream task to evaluate the quality of the semantic representation learned by our method.
The higher the user identification accuracy, the better the quality of the learned representation.

\paragraph{Datasets}
We evaluated on the EgoBody dataset considering that it provided user labels for every recorded sequence.
We noticed that the default training and test sets provided in the original paper~\cite{zhang2022egobody} contain no overlapping users and thus cannot be directly used for evaluating user identification.
To address this problem, we opted to only use the default test set and split it into two halves with each half containing half of the data from $12$ users.
We used one half for training and the other half for testing.

\paragraph{Baselines}
We compared our method with the baseline methods as described in \autoref{sec:eval_settings}.
We also provided a ``Chance'' baseline to help understand the results with more context.
In addition, to evaluate the effectiveness of using hand-head forecasting as an auxiliary training task, we further used the ablated version of not using hand-head forecasting as a baseline method.
Furthermore, we employed our method to learn representations \textit{only} from hand or head movements and used them as the baselines to compare with our hand-head joint representations.

\paragraph{Procedure}
We followed common practice of evaluating representation learning methods~\cite{zhang24mouse2vec, zhang24dismouse, preechakul2022diffusion} to first calculate the semantic representations of hand-head behaviours using the pre-trained representation methods and then apply these representations as pre-computed features to train and test a classifier for user identification.
The classifier contains a linear layer and a Softmax activation function to compute the probability of each user.
We trained the classifier using the Adam optimiser with a learning rate of $1e-5$ and a batch size of $64$ for a total of $60$ epochs for different representation methods, respectively.

\paragraph{Results} 
\autoref{tab:downstream} shows the user identification accuracies of different representation methods on the EgoBody dataset.
It can be seen in the table that our method can achieve superior performance over other methods in terms of identification accuracy ($29.8\%$ vs. $28.0\%$).
Using a paired Wilcoxon signed-rank test we confirmed that the differences between our method and other methods are statistically significant ($p<0.01$).
We also find that our method can achieve higher identification accuracy than the ablated version of not using hand-head forecasting ($29.8\%$ vs. $29.4\%$), validating the usefulness of hand-head forecasting as an auxiliary training task to refine the semantic representation.
Furthermore, we observed that the hand-head joint representations learned by our method significantly outperform the representations learned only from hand trajectories ($29.8\%$ vs. $18.0\%$) or head orientations ($29.8\%$ vs. $25.7\%$), demonstrating the superiority of our joint representations.
The above results exhibit the effectiveness of our method as a generic feature extractor for the application of user identification.

\subsubsection{Activity Recognition} \label{sec:activity_recognition}

Activity recognition is an important task in the area of human-centred computing and has significant relevance for many XR scenarios such as adaptive virtual environment design~\cite{hadnett2019effect}, low-latency predictive interfaces~\cite{david2021towards, keshava2020decoding}, and human-aware intelligent systems~\cite{vortmann2020attention}.
It is well-known that user activities can be recognised from their hand and head movement patterns~\cite{hu2022ehtask, ohn2014head}.
As such, activity recognition serves as a particularly relevant sample downstream task to further evaluate the semantic representation learned by our method.
The higher the activity recognition accuracy, the better the quality of the learned representation.

\paragraph{Datasets}
We used the EgoBody and ADT datasets for evaluation given that they provided activity labels for each collected sequence.
For the EgoBody dataset, we used the default training and test sets provided by the authors~\cite{zhang2022egobody} (see \autoref{sec:datasets}) and followed prior work~\cite{hu24_pose2gaze} to evaluate for three activities, i.e. \textit{chat}, \textit{learn}, and \textit{teach}, that have the most recordings.
For the ADT dataset, we used $24$ sequences for training and $10$ sequences for testing to evaluate for three activities, i.e. \textit{decoration}, \textit{meal}, and \textit{work}, following prior work~\cite{hu24_pose2gaze}.

\paragraph{Baselines}
We employed the same baselines as used for user identification (see \autoref{sec:user_identification}).

\paragraph{Procedure}
We followed the same procedure as used in \autoref{sec:user_identification} to first calculate the semantic representations and then train and test a linear classifier for activity recognition.

\paragraph{Results}
\autoref{tab:downstream} shows the activity recognition accuracies of different representation methods on the EgoBody and ADT datasets.
We can see that our method consistently outperforms other methods in terms of recognition accuracy on both the EgoBody ($55.7\%$ vs. $50.5\%$) and ADT ($63.9\%$ vs. $62.7\%$) datasets and the differences between our method and other methods are statistically significant (paired Wilcoxon signed-rank test, $p<0.01$).
We also noticed that our method achieves higher recognition accuracies than the ablated version of not using hand-head forecasting on both EgoBody ($55.7\%$ vs. $54.7\%$) and ADT ($63.9\%$ vs. $63.1\%$), demonstrating the effectiveness of using hand-head forecasting as an auxiliary training task to refine the semantic representation.
In addition, we confirmed that our joint representations can achieve higher performance than the representations learned only from hand or head movements 
on both EgoBody and ADT, proving the superiority of our joint representations.
These results illustrate the usefulness of our method as a reusable feature extractor for the task of activity recognition.

%% file: sections/discussion.tex
\section{Discussion}

\subsection{On Performance}
In this work, we proposed \methodName~-- the first self-supervised method to learn generalisable joint representations of human hand and head movements in extended reality.
We evaluated our method in the challenging out-of-domain setting, i.e., we trained it on one XR dataset and tested it on two unseen datasets across different users, activities, and XR environments.
We showed that our method significantly outperforms other methods in hand and head reconstruction performance, achieving improvements of up to 74.1\% (see \autoref{tab:results}).
We also demonstrated that our method has strong generalisation capability for different users, activities, and XR environments (see \autoref{tab:results}). 
We further evaluated the learned semantic representations on two sample downstream tasks relevant to XR applications: understanding users and their interaction context.
As shown in \autoref{tab:downstream}, the semantic representations learned by our method consistently outperform those learned by other methods across different tasks and different datasets, achieving an improvement of up to 5.2\% (55.7\% vs. 50.5\% on EgoBody) in terms of activity recognition accuracy.
Taken together, these results are highly promising and, for the first time, demonstrate the feasibility of learning generalisable joint representations of hand and head movements in a self-supervised fashion.
That is, without requiring additional human annotations that are usually costly, cumbersome, and time-consuming to collect~\cite{chu2023wordgesture}.
As such, our work opens up an exciting new research direction for hand-head behaviour modelling in extended reality.

\subsection{On the Method}
Our method combines a GCN-based and a diffusion-based encoder to learn joint semantic and stochastic embeddings from hand-head signals, respectively, and a diffusion-based decoder to reconstruct the original signals from the learned embeddings.
We used a graph convolutional network as our semantic encoder to extract features from hand and head movements owing to its good performance in fusing information from different modalities~\cite{hu24gazemotion, yan24gazemodiff, hu24hoimotion}.
We used a DDIM model because it can explicitly encode the original signals' stochastic variations and can reconstruct the original data with high quality from the semantic and stochastic embeddings.
The results summarised in \autoref{tab:results} showed that both the learned semantic ($E_{sem}$) and stochastic ($E_{sto}$) representations contribute significantly to the reconstruction performance of our method while $E_{sem}$ has higher influence on reconstruction than $E_{sto}$, validating that more critical information of hand-head signals is encoded in $E_{sem}$.

Even more interestingly, by disentangling the hand-head joint representations into a semantic and a stochastic part, our method enables novel use cases with significant potential for XR applications.
Specifically, the semantic representation can be used to analyse hand-head behaviours in XR by identifying interpretable hand-head clusters within the original signals (see \autoref{fig:clusters}) while the stochastic representation can be used to generate hand-head movements with controllable randomness (see \autoref{fig:generation}).
Given these results, our method contributes to the explainability of human hand and head behaviours in XR and can serve as an analysis or generation tool to facilitate future research in human hand-head behaviour modelling and help develop future human-aware intelligent XR systems.
Furthermore, our method also provides meaningful insights into learning disentangled representations of human behaviours in extended reality.

\subsection{On Hand-head Joint Representations}
In this work, we pioneered a new method for learning joint representations of human hand and head movements in XR, while prior works on hand and head behaviour modelling typically only explored a single modality~\cite{belardinelli2022intention, bachynskyi2020dynamics, hu2019sgaze, hu2020dgaze}.
We chose to model hand and head behaviours jointly based on the insight that human hand and head movements are closely coordinated with each other in most daily activities~\cite{hu24_pose2gaze, emery2021openneeds}.
Joint modelling of hand and head behaviours offers many benefits for practical applications in extended reality:
First, it is easier to interpret the semantics behind human behaviours when considering both human hand and head movements than only using a single modality.
For example, in \autoref{fig:clusters}, it would be much more difficult to understand the users' behaviours if hand trajectories or head orientations were removed.
In addition, hand-head joint representations make it possible to simultaneously generate hand and head movements that are coordinated with each other, as illustrated in \autoref{fig:generation}.
Furthermore, joint representations of hand-head movements can capture more useful information than the representations learned from only hand or head signals and can achieve better performance when applied to downstream tasks (\autoref{tab:downstream}).
Given these benefits, it is highly promising to continue working in this direction to develop more powerful methods for learning hand-head joint representations.

\subsection{Limitations and Future Work}
Despite all these advances, we also identified several limitations that we plan to address in future work.
First, to learn generalisable representations across different XR environments, we used relative coordinates with the origin set to the position of the head. 
However, in doing so we neglected the translational movement of the head and only encoded the head's rotational movement in our representation.
In the future, we plan to explore how to also include translational components of the head movements into the learned latent embeddings.
In addition, we trained our method only on the EgoBody dataset and directly tested it on the ADT and GIMO datasets.
In future work we would like to train and test our method on more datasets to further evaluate its generalisability.
Finally, in the future, we would also like to explore more applications of our hand-head joint representations to assist XR researchers in analysing and understanding human behaviours.

%% file: sections/conclusion.tex
\section{Conclusion}

In this work, we proposed a novel self-supervised method for learning generalisable joint representations of human hand and head movements in extended reality that first uses a GCN-based semantic encoder and a DDIM-based stochastic encoder to learn the semantic and stochastic representations, respectively, and then applies a DDIM-based decoder to reconstruct the original hand-head data.
Through extensive experiments on three public XR datasets for reconstructing the original hand-head signals, we demonstrated that our method significantly outperformed other methods by a large margin and can be generalised to different users, activities, and XR environments.
We further show that the representations learned by our method can be used to analyse hand-head behaviours, generate variable hand-head data, and serve as pre-computed features for two practical sample downstream tasks.
As such, our results underline the potential of applying self-supervised methods for jointly modelling human hand-head behaviours in extended reality and guide future work on this promising research direction.